\documentclass[11pt,a4paper]{article}
\usepackage[hyperref]{acl2020}
\usepackage{times}
\usepackage{latexsym}

\usepackage{microtype}

\usepackage{balance}

\aclfinalcopy % Uncomment this line for the final submission
\title{Language Model Metrics and Procrustes Analysis\\ for Improved Vector Transformation of NLP Embeddings}

\author{
  Thomas Conley \\
  University of Colorado\\Colorado Springs\\
  1420 Austin Bluffs Pkwy\\
  Colorado Springs, CO, USA\\
  {\tt tconley@uccs.edu} \\\And
  Jugal Kalita \\
  University of Colorado\\Colorado Springs\\
  1420 Austin Bluffs Pkwy\\
  Colorado Springs, CO, USA\\
  {\tt jkalita@uccs.edu} 
  }

\date{}

\usepackage{graphicx} %Loading the package
\graphicspath{
    {./images/}
    {/Users/tconley/_school/projects/ICSC2019/images/}
    {/Users/tconley/_school/src/cs5870-embeddings/logs/}
    {../}
    {/Users/tconley/_school/src/2020/word2vec/plots/}
    }
\usepackage{balance}
\usepackage{amsmath}

\begin{document}
\maketitle
\begin{abstract}
	Artificial Neural networks are mathematical models at their core. This truism presents some fundamental difficulty when networks are tasked with Natural Language Processing. A key problem lies in measuring the similarity or distance among vectors in NLP embedding space, since the mathematical concept of distance does not always agree with the linguistic concept. We suggest that the best way to measure linguistic distance among vectors is by employing the Language Model (LM) that created them.  We introduce Language Model Distance (LMD) for measuring accuracy of vector transformations based on the Distributional Hypothesis (\emph{LMD\_Accuracy}). We show the efficacy of this metric by applying it to a simple neural network learning the Procrustes algorithm for bilingual word mapping.  
 	\end{abstract}

\section{Introduction}

  The Distributional Hypothesis \cite{firth1961papers} inspired the development of  embeddings that capture the meaning of language based on how words co-occur with each other \cite{mikolov2013distributed}. 
  Natural Language Processing relies heavily on these high dimensional vectors to represent words, phrases, sentences or documents, in a form that can be processed by deep neural networks which were originally designed for tasks related to computer vision. 
  Input embeddings are transformed by network layers into output vectors which represent  solutions to many NLP tasks \cite{ruder2019survey}.     

  In order to learn these transformations, a network must be able to calculate the difference between predicted vectors and actual word vectors.  
  This distance calculation is a crucial part of measuring loss, and performing back-propagation. 
  These core functions of neural networks have primarily relied on mathematical processes without regard to linguistic principles. We demonstrate that NLP embedding transformation is better measured using linguistic similarity functions rooted in knowledge of languages rather than concepts such as Euclidean or angular distance, which assumes vectors to be ``physical" objects.  
  
  \subsection{Procrustes Analysis}
  
  Matrix transformation of vector spaces has been accomplished using Generalized Procrustes Analysis (GPA), ever since a computationally viable solution was devised \cite{gower1975generalized}. In particular, GPA has been used to great effect in geo-spatial shape manipulation \cite{Duta2015procrustes,crosilla2019orthogonal} and qualitative data analysis \cite{crusty-food2016}.   

  Shapes are represented by a series of landmark points in 2 or 3 dimensions. And in survey research, qualitative opinion data is represented by a Likert scale \cite{likert1932technique}, occupying low dimensional space.  In both cases, the vector spaces must be realigned and resized for meaningful comparison.  Although these fields seem to differ, they use data structures that share characteristics with Natural Language Processing.
 
  The orthogonal Procrustes algorithm produces an optimal transformation matrix $R$ for mapping one vector space to another and appears to be useful in converting vector spaces for NLP tasks such as bilingual word mapping \cite{kementchedjhieva2018generalizing}.       

  \subsection{Procrustes Analysis for NLP tasks}

  Can a neural network learn to do Procrustes transformation?  The answer, yes, should be non-controversial, since every neural network performs tensor transformation of input to output.  However, tasks which require nuanced understanding of the meaning of words, such bilingual word mapping, are particularly difficult.  Although there is some success when massive amounts of text are available for training, the problem is more acute when resources for learning are scarce, as in machine translation of under resourced languages.  

  The difficulty with vector transformations in NLP is based on the nature of the data. NLP transformations by neural networks use distance measurements designed to work in $L_p\;space$. This implies numerical data.  We show that such calculations of distance and accuracy are not as effective as measurements based on language models.

  \subsection{Image data and language data}

  We consider image data as raw data with physical dimensionality where, each dimension in a vector can be considered similar in measurement and meaning.  As such, this data occupies $L_p\;space$; where vectors can be added together or multiplied by scalars without loss of their inherent meaning. For example, a vector representing a pixel is measured the same way, and has the same meaning, regardless of where it is in the image.

  Thus, distance measurement among image vectors can use $L_p\;norm$ or trigonometric calculations such as cosine distance.  One specific kind of euclidean distance measurement is called \emph{Procrustes Distance} and is the basis of Procrustes Analysis \cite{crosilla2019orthogonal}.

% axioms
  In NLP, distance measurement is less meaningful when it is based on Euclidean axioms rather than linguistic principles.  Distance is the basis of error calculations and back-propagation, and so, the ability to calculate the derivative of these functions is essential for classic stochastic gradient descent (SGD) which is employed by neural networks today.  Although there has been some research in non-differentiable losses \cite{engilberge2019sodeep} the mathematical requirements for these functions are not always suitable for NLP.  

  % In raw data, each dimension is considered to have equal validity or importance in distance calculations, even though the position of the dimension in a vector make a difference in importance.  For instance in a vector of pixel data representing an image, pixels in the center of the image vector may be more important than peripheral pixels. 

  As opposed to raw data, feature data consists of vectors in which each dimension may have disparate meaning and measurement.  Feature data does not exist in $L_p\;space$, and therefore measures of distance that rely on $L_p\;norm$ or trigonometric calculations may not be meaningful.  We consider NLP embeddings to be feature data, although they share some  characteristics with raw data.  
 
  % We use the term NLP vectors or embeddings to imply that these data may include words, phrases, sentences or documents. We emphasize that neural network metrics function on the vectors themselves regardless of the constructs they represent, e.g., the mathematics of the vectors, not the linguistics.    

\section{Language Models and Data}

  As in raw data, NLP vectors dimensions typically share values that are treated similarly and are thus undifferentiated in a sense.  This seems to contradict the assertion that each NLP embedding dimension has a specific unique meaning like feature data.  Instead, the meaning of a dimension is more like probability, representing how often a word is used with a particular meaning, rather than the actual meaning of the word.  
  
  Vectors with dimensions that differ in meaning, as in NLP embeddings, cannot be used with typical spatial measurements such as $L_p\;norm$ and $cosine\;distance$. We contend that NLP vector distance can best be measured by the language models which represent the vectors.  Therefore, we seek to replace mathematic calculations with predictions from language models.  We simply rely on the language model itself to provide a distance measurement for our custom metric. 

  % We acknowledge that such methods may not be differentiable.  

  In this research, we use the Word2Vec model \cite{word2vec} to produce a custom bilingual word mapping dataset.  This dataset, combined with the GenSim model of keyed vectors \cite{gensim}, provides a distributional distance measurement based on word movers distance \cite{wordmoversdistance}.  

  % It is important to note that the models we employ, English and Spanish, are trained separately and do not share knowledge of the other.   
  % We choose 1000 Spanish to English word pairs as a test set for this model and for our custom metric.

  Our neural network is a simple Multilayer Perceptron (MLP) which accepts Spanish word vectors as input and predicts English word vectors. This simple model was chosen because it is analogous to any layer found in innumerable, more complex, neural networks. Showing improved efficacy in this model should demonstrate improvement in any NLP task.

  \begin{figure}[ht]
  \includegraphics[width=\columnwidth]{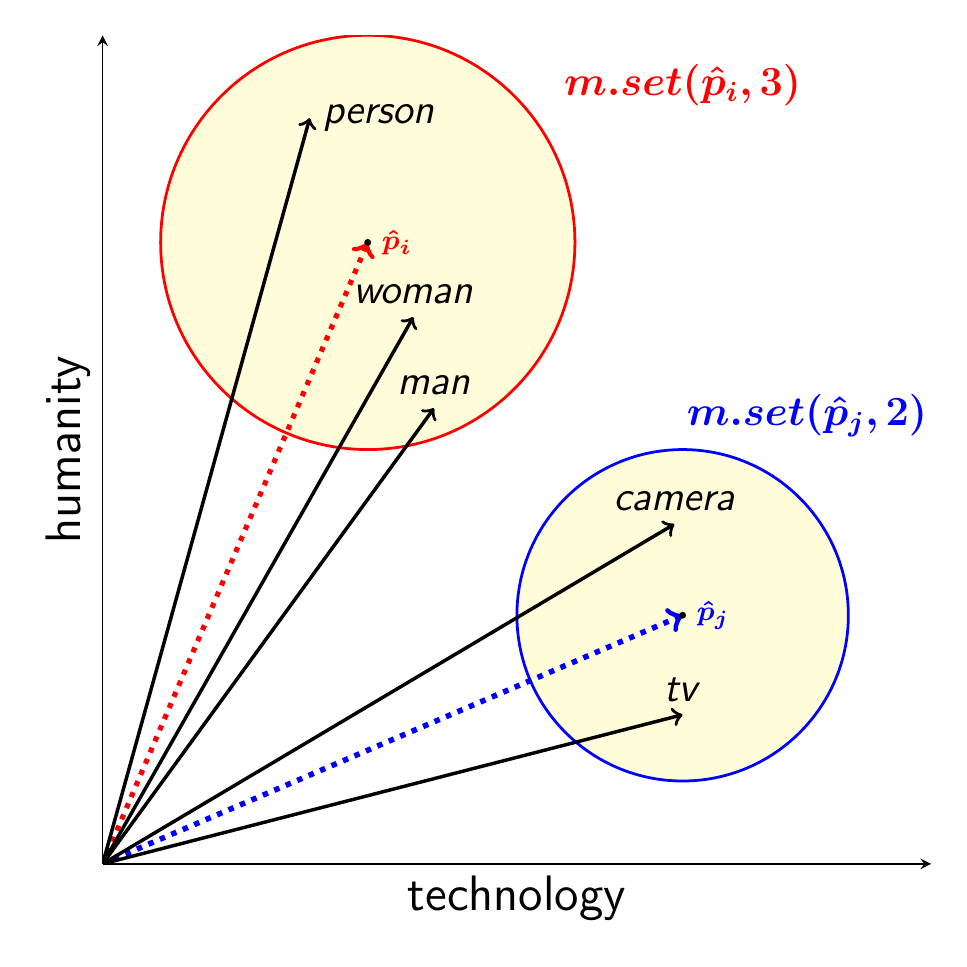}
  \caption{Illustration of the Distributional Hypothesis and Language Model Distance.  The accuracy of predicted vectors $\hat{p_i}$ and $\hat{p_j}$, is based on membership in the set of $k=2$ or $k=3$ neighbors. }
  \label{fig:person}
  \end{figure}

\section{Language Model Distance}

  An exact measurement of equality is not possible for high-dimensional NLP embeddings.  Embeddings of several hundred dimensions, and one-hot encoded vectors on the order of tens of thousands of dimensions, are particularly difficult to measure.  
  % Metrics such as $L_p\: norm$ or Cosine Similarity may not adequately capture the meaning of words, phrases, or documents due to the sparsity of high-dimensional space.  

\begin{equation} \label{eq:distacc}
  \small
      LMD(\hat{p},t,m,k)= 
  \begin{cases}
      True, & \text{if}\; t \in m.set(\hat{p},k)\\
      False, & \text{otherwise}
  \end{cases}
  \end{equation}

  Instead, we suggest that the true measure of NLP vector distance is best provided by the model which defines the vectors. We present a family of metrics, Language Model Distance (LMD), which calculates distance and equality among NLP vectors by using the language model itself.  $LMD$ is defined as in Equation \ref{eq:distacc} where the distance between predicted vector $\hat{p}$ and known truth vector $t$, is provided by model $m$, given neighbor threshold $k$. 

  The distance measure is binary because it is based on set inclusion, and not physical or Euclidean distance. Thus, $LMD$ can be used as a measure of accuracy, and records a true positive when $t$ is within the neighborhood of the predicted vector ($t\in m.set(\hat{p},k)$). 

  %Note that the distance measure ($LMD(\hat{p},t,m,k)$) is binary, based on set inclusion and not Euclidean distance, thus it can be used in measures of accuracy.

  \subsection{Measuring Accuracy with Language Model Distance}

    Figure \ref{fig:person} illustrates the distributional hypothesis by showing a simple clustering along 2 non-numeric dimensions.  The circles represent neighborhoods $m.set(\hat{p},k=2)$ and $m.set(\hat{p},k=3)$.  Note that the predicted vectors ($\hat{p}$) have no words directly associated with them, because no exact match is possible for floating point numeric vectors.  

    Thus we say that $LMD\_Accuracy(k)$ measures a positive result when truth vector ($t$) is within the $k$ sized neighborhood of the predicted vector ($t\in m.set(\hat{p},k)$).  For example, $LMD\_Accuracy(3)$ measures the percentage of times that the true word answer was among the top 3 closest predicted words. 
  
    % In reality language models express NLP spaces with many more dimensions than shown in Figure \ref{fig:person}.  Viewing clusters of word NLP vectors shows that the dimensions have meaning, although it is not always possible to discern the actual meaning.

    Distributional distance functions can be used in neural network metrics, loss, or activation functions, or used directly in similarity computation.  However, inserting external language models into neural networks can be difficult as these networks are firmly rooted in mathematics which is not compatible with linguistic processes. 

\begin{figure}[ht] 
    \setlength{\fboxsep}{6pt}
    \fbox{\includegraphics[width=.9\columnwidth]{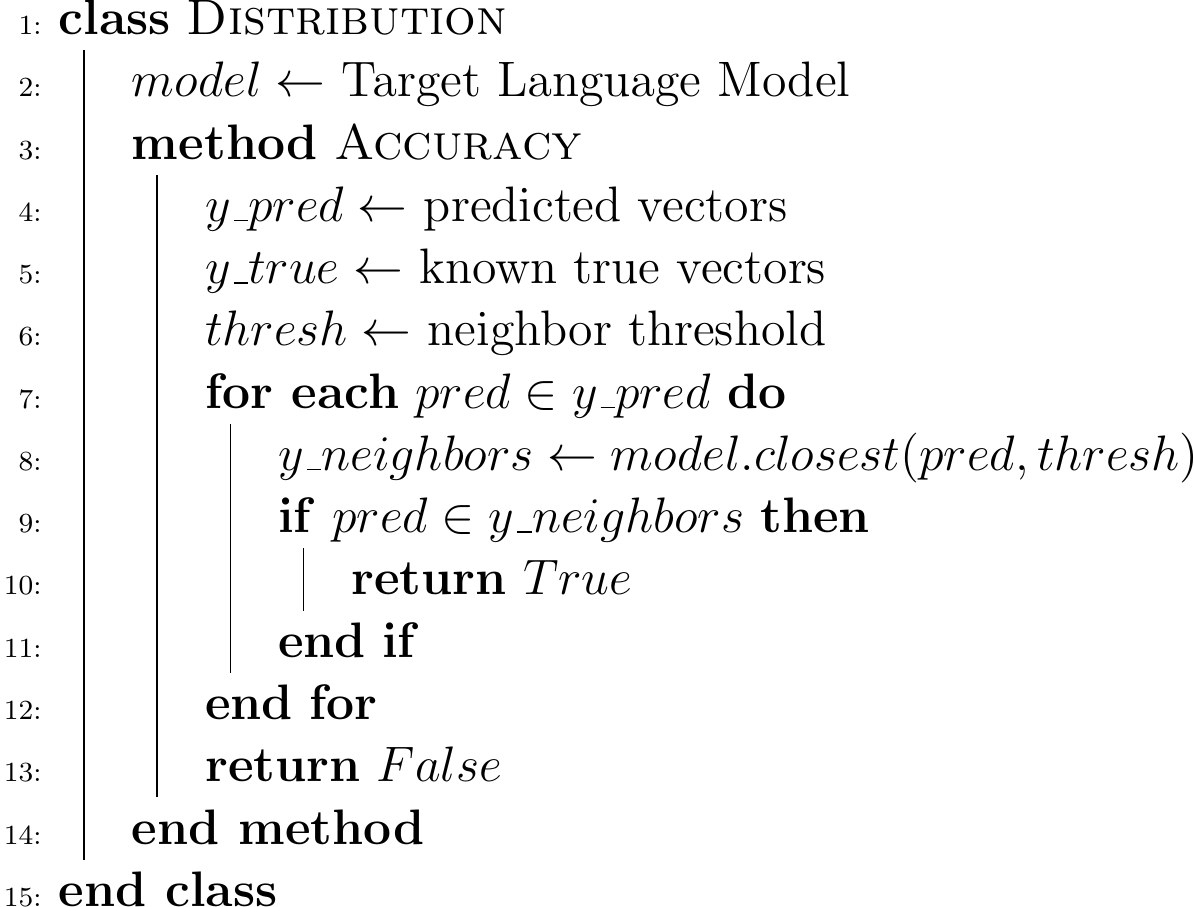}}
    \caption{Implementation of Distributional Accuracy based on Language Model Distance.  A static language model (line 2) allows linguistic functionality to be included in purely mathematical models.}
    \label{fig:alg}
    \end{figure}

    We solve these difficulties by defining a simple class shown in Figure \ref{fig:alg}. By including the language model as a static member of the class, methods of the class may be used as network internal functions with access to external language models.

 	% We define the term \emph{Distributional Accuracy} as a measure of proximity among NLP vectors in high dimensional space based on the Distributional Theory\cite{mikolov2013}. The measure of accuracy is simple but the distance measure must be provided by a model, A simple accuracy as proximal measure of closeness based on a model that embodies the Distributional Hypothesis.

\section{Learning Orthogonal Procrustes Analysis}

% 2or 3
  The  Orthogonal Procrustes Algorithm is a process for finding the optimal mapping of one set of vectors to another. Typically, the vectors represent points in 2 or 3 dimensional space, for image processing, or they represent qualitative data measured in few dimensions \cite{crusty-food2016}. After resizing and repositioning of vectors, an optimal rotation matrix $R$ is produced by a method similar to singular value decomposition.  

  This classic approach to vector transformation has been explored as a solution for some NLP tasks \cite{sen2019multilingual,kim2019pivot}.  Therefore we ask: Can a neural network be trained to perform the same optimal transformation for NLP embeddings which occupy a much higher dimensional space?    

%
  % The first step in the algorithm, known as translation, calculates the mean of all vector dimensions and repositions the vectors around a the common origin. Second, the vectors are uniformly scaled by reducing the root mean squared error of the all dimensions. Finally, an optimal rotation matrix $R$ is calculated using 

  Our task is to train a simple MLP to learn the optimal mapping $R$, between two disparate vector spaces representing a bilingual dataset. We measure the success of this task using $LMD$ as the basis for accuracy as in Figure \ref{fig:alg} and Equation \ref{eq:distacc}. 

  We create two separate language models from a parallel corpus of European Parliament translations, the so called EuroParl dataset \cite{koehn2005europarl}. We use the Word2Vec model in continuous bag-of-words (CBOW) mode \cite{mikolov2013distributed} to build two separate distributions.  By using a bilingual corpus, and training language models separately,  we ensure that the models share a common domain, but the vector spaces remain separate.  For training, we then map word vectors from one distribution to the other, using a set of 1000 most common words pairs, obtained from from a language learning website\footnote{http://www.englishnspanish.com}.

  \subsection{Results}
\begin{figure}[ht]
      \includegraphics[width=\columnwidth]{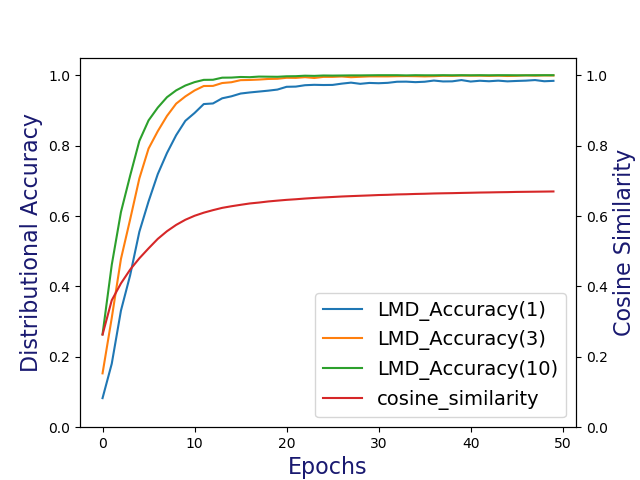}
      \caption{Results of Learning Orthogonal Procrustes Analysis showing a better measure of exact matches with \emph{LMD\_Accuracy} than with \emph{cosine similarity}.}
      \label{fig:classic}
    \end{figure}

    Our results show that Orthogonal Procrustes Analysis can be learned for  multilingual mapping of word vectors.  Furthermore, Figure \ref{fig:classic} demonstrates that LMD is effective as a basis for measuring the accuracy of this task.  

    Figure \ref{fig:classic} indicates that \emph{LMD\_Accuracy} is better at measuring similarity in NLP embeddings than \emph{cosine similarity}. In this plot, \emph{LMD\_Accuracy(1)} indicates that the model exactly predicted the correct word in the output language.  When \emph{LMD\_Accuracy(1)} is near 100\% the value of \emph{cosine similarity} should be near $1$ which would indicate an exact match.  The fact that \emph{cosine similarity} cannot measure this exact match shows a weakness in this purely mathematical measurement compared with our language model-based measurement. 

  \section{Learning General Procrustes Analysis}

    To further test, we try to learn General Procrustes Analysis; a much harder task because it requires the network to generalize.

    We have just shown that a simple neural network can learn to transform vectors.  This is non-controversial since all neural networks perform this task at every layer.  However, not all networks are able to generalize. Using the same network configuration as before, we now evaluate embeddings that we have not seen in training, as is common. This is equivalent to learning the \emph{Generalized Procrustes Algorithm}. 

    \begin{figure}[ht]
    \includegraphics[width=\columnwidth]{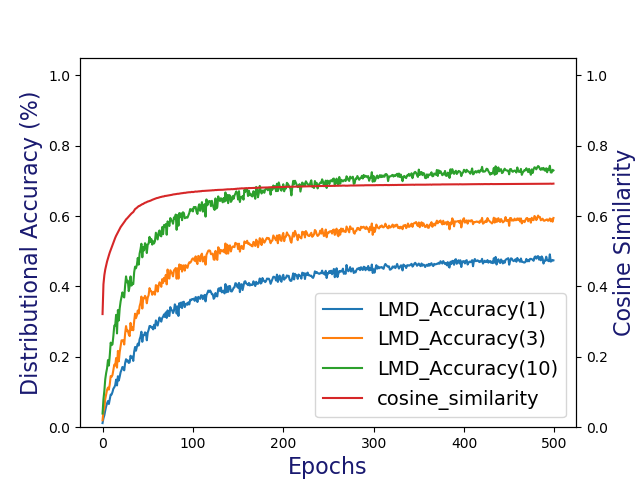}
    \caption{Results of Learning General Procrustes Analysis showing a comparable measure of exact matches between \emph{LMD\_Accuracy} and \emph{cosine similarity}, when generalization is required}.
    \label{fig:general}
    \end{figure}

    Results in Figure \ref{fig:general} show that \emph{LMD\_Accuracy} is more like cosine distance when generalization is required.  Note that we use  \emph{LMD\_Accuracy} only for metrics.  This model uses $cosine\;similarity$ for error calculation and back-propagation. We conclude that such $L_p\;norm$ measurements can only drive generalization as far as they are able to measure accuracy.

    The local variation in \emph{LMD\_Accuracy}, evident in Figure \ref{fig:general}, may be significant as it may make determining the derivative of the function difficult.  The derivative of \emph{LMD\_Accuracy} must be worked out before it can be incorporated into a loss function and be used in back-propagation. The overall shape of the curve, despite irregularities is encouraging as the slope may be computed using ordinary least squares in a calculation of rolling regression, or by other numerical methods.  

    \section{Conclusion}
    We suggest that language model metrics described here may be incorporated directly into activation and loss functions, and may be used as an error measurement for back-propagation.  We suggest this basic enhancement would improve the Generalized Procrustes Algorithm and other NLP processing in general.  This is left for future work.

  % \subsection{Conclusion}

  %   Neural networks have always used calculations in $L_p\: space$ for activation and loss functions.  Thus they are able to learn many functions easily depending on the kind of vector space on which they operate.

  %   We believe that language model functionality can be directly incorporated into neural network processing at the core.  We are designing activation and loss functions that will allow language knowledge to drive stochastic gradient descent and back-propagation.    

% \nocite{stegmann2002brief}
% \nocite{klingenberg2015}
% \nocite {crusty-food2016}

\balance  
\bibliographystyle{acl_natbib}
\bibliography{icon2020}

\end{document}